\pdfoutput=1

\documentclass[11pt]{article}

\usepackage{acl}

\usepackage{times}
\usepackage{latexsym}

\usepackage[T1]{fontenc}

\usepackage[utf8]{inputenc}
\usepackage{graphicx}
\usepackage{algorithm}
\usepackage{algorithmic}
\usepackage{microtype}
\usepackage{booktabs}
\usepackage{amsmath}
\usepackage{listings}
\usepackage{xcolor}

\definecolor{codegreen}{rgb}{0,0.6,0}
\definecolor{codegray}{rgb}{0.5,0.5,0.5}
\definecolor{codepurple}{rgb}{0.58,0,0.82}
\definecolor{backcolour}{rgb}{0.95,0.95,0.92}

\lstdefinestyle{mystyle}{
    backgroundcolor=\color{backcolour},   
    commentstyle=\color{codegreen},
    keywordstyle=\color{magenta},
    numberstyle=\tiny\color{codegray},
    stringstyle=\color{codepurple},
    basicstyle=\ttfamily\footnotesize,
    breakatwhitespace=false,         
    breaklines=true,                 
    captionpos=b,                    
    keepspaces=true,                 
    numbers=left,                    
    numbersep=5pt,                  
    showspaces=false,                
    showstringspaces=false,
    showtabs=false,                  
    tabsize=2
}

\title{\textit{DiffChat}: Learning to Chat with Text-to-Image Synthesis Models for Interactive Image Creation}

\author{Jiapeng Wang$^{1}$\thanks{\ \ Contribution during internship at Alibaba Group.}, Chengyu Wang$^{2}$\thanks{\ \ Co-corresponding authors.}, Tingfeng Cao$^{1}$, Jun Huang$^{2}$, Lianwen Jin$^{1}$\footnotemark[2]\\
  $^{1}$ South China University of Technology, China \\
  $^{2}$ Alibaba Group, China \\
  \texttt{\{eejpwang, setingfengcao\}@mail.scut.edu.cn, eelwjin@scut.edu.cn}\\
  \texttt{\{chengyu.wcy, huangjun.hj\}@alibaba-inc.com}
}

\begin{document}
\maketitle
\begin{abstract}
We present \textit{DiffChat}, a novel method to align Large Language Models (LLMs) to ``chat'' with prompt-as-input 
Text-to-Image Synthesis (TIS)
models (e.g., Stable Diffusion)  for interactive image creation. 
Given a raw prompt/image and a user-specified instruction,  \textit{DiffChat} can effectively make appropriate modifications and generate the target prompt, which can be leveraged to create the target image of high quality. To achieve this, we first collect an instruction-following prompt engineering dataset named InstructPE for the supervised training of \textit{DiffChat}.
Next, we propose a reinforcement learning framework with the feedback of three core criteria for image creation, i.e., aesthetics, user preference and content integrity. 
It involves an 
action-space dynamic modification technique to obtain more relevant positive samples and harder negative samples during the off-policy sampling. 
Content integrity is also introduced into the value estimation function for further improvement of produced images. 
Our method can exhibit superior performance than baseline models and strong competitors based on both automatic and human evaluations, which fully demonstrates its effectiveness.
\footnote{InstructPE is available at EasyNLP \cite{DBLP:conf/emnlp/WangQZLLWWHL22}. URL:~\url{https://github.com/alibaba/EasyNLP}}

\end{abstract}

\section{Introduction}

\begin{figure}[t]
\centering
\includegraphics[width=0.49\textwidth]{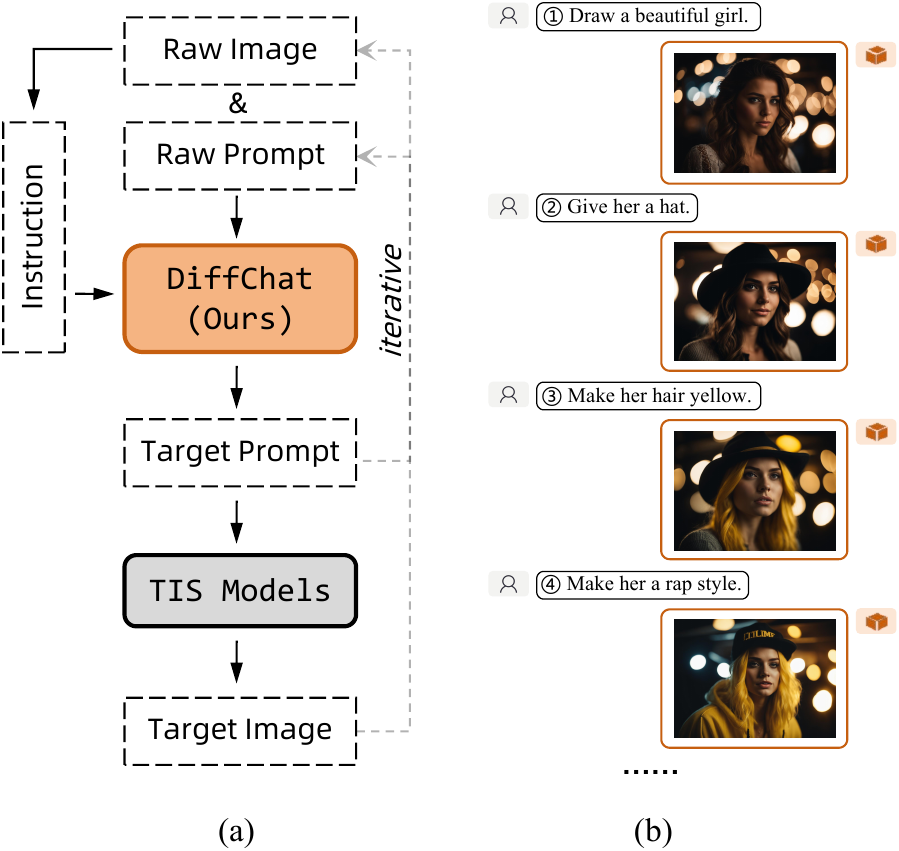}
\caption{(a) The pipeline of our \textit{DiffChat} collaborating with off-the-shelf TIS models for interactive image iteration.
(b) A simple example of \textit{DiffChat} following instructions to interact with TIS models (Stable Diffusion XL here) for interactive image creation. Note that~\textit{DiffChat} is capable of automatic prompt refinement and re-writing through ``chats'' and can be applied to a variety of TIS models.
} 
\label{fengmian}
\end{figure}

In recent years, large-scale deep generative models have emerged as powerful tools for generating contents across various modalities.
One of the most remarkably developed and extensively adopted applications is Text-to-Image Synthesis (TIS), which aims to create realistic images with texts as inputs (\textit{prompts}). Large pre-trained TIS models \cite{dalle,dalle2,stablediffusion,imagen,controlnet,BlendedLD} have achieved significant improvement to enable users to create images of unprecedentedly high quality, even without art expertise.

However, for non-experts, coming up with appropriate and accurate prompts required by TIS models is by no means an easy task. 
Different language expressions with the same semantics or slightly minor revisions can often result in multiple variations of image generation, which means it is full of uncertainty to write such prompts that meet the user's requirements~\cite{uncertain1,uncertain2}.
Furthermore, when non-experts wish to create images with specific needs,
they often need to iteratively conduct uncertainty trials and errors for prompt refinement, leading to significant losses of time and computing resources. The capabilities of TIS models are also under-utilized in this case, due to the poorly optimized prompts.

To address these issues, in this paper, we propose a novel framework named \textit{DiffChat}, which can follow user-specified instructions to interact with TIS models for image creation, as shown in Fig.~\ref{fengmian}.
It avoids the tedious attempts of prompt crafting and re-writing mentioned above, making users feel as simple as ``chatting'' with these TIS models.
Specifically, we first collect an \textbf{Instruct}ion-following \textbf{P}rompt \textbf{E}ngineering dataset named \textbf{InstructPE} using an automatic data collection pipeline based on existing datasets and AI models. 
Next, we utilize InstructPE to align off-the-shelf LLMs to adapt to our task with supervised fine-tuning. 
Finally, we propose an enhanced reinforcement learning framework with three user-concerned criteria for image creation:
(1) \textit{\textbf{A}esthetics} which represents the aesthetic evaluation of created images;
(2) \textit{\textbf{P}reference} that indicates the user's preference for specified images relative to other ones;
and (3) \textit{\textbf{C}ontent integrity} to evaluate whether the creations contain core contents as complete as possible.
Thus, \textit{DiffChat} can be trained to pursue positive  \textit{\textbf{APC}} feedback without any manual labeling.
To further enhance the sample quality during training, 
we also propose an improved sampling strategy based on Action-space Dynamic Modification (ADM). For positive samples, we restrict the generation of tokens with low information quantity to improve the overall quality; and for negative samples, we also partially mask or replace the key information to simulate hard samples, allowing the model to fully learn from these errors.
Additionally, we develop the Value estimation function of vanilla PPO \cite{ppo} method with the consideration of Content Integrity (VCI), to achieve a more accurate perception of the current state during optimization. 

By using prompts as intermediaries, \textit{DiffChat}  allows users to easily interact and collaborate with TIS models for image creation through chatting.
It is worth noting that, recent related research works~\cite{InteractiveImageManipulation,BATINeT,instructpix2pix,couairon2023diffedit,zhang2023hive,emu,gill} mainly focus on the model structure design. They typically adopt instructions, images, and additional information as inputs to generate edited images using an end-to-end network.
Different from these works, our method pays more attention to the preliminary automatic prompt writing procedure, which can be used in conjunction with these methods or various widely-used TIS models such as the series of Stable Diffusion (SD) models.
It does not need to re-train with the development of TIS models
to make extra costs, 
with its user-friendliness and generalization abilities fully manifested.

Experimental results based on both automatic and human evaluations demonstrate that our method exhibits greater performance than baseline models and competitors.
In summary, the main contributions of this paper are listed as follows:
\begin{itemize}
\item We propose \textit{DiffChat} to collaborate with TIS models for interactive image creation. It is easy to use and applicable to a wide range of TIS models. Surpassing baselines and competitors also indicates its effectiveness.

\item We release a new prompt engineering dataset named~\textbf{InstructPE} with 234,786 train and 5,582 test samples for supervised fine-tuning. We further conduct feedback on aesthetics, preference, and content integrity during reinforcement learning.  ADM and VCI are also introduced for improved off-policy sampling and state value estimation, respectively.

\item The public availability of InstructPE
is expected to 
 greatly promote future research on TIS based on instruction-based user-agent interaction.
\end{itemize}

\section{Related Work}
\subsection{Text-to-Image Synthesis (TIS) Models}
TIS is a multi-modal task involving the generation of images based on textual conditioning. 
In early years, prevalent research works for TIS primarily relied on the concept of generative adversarial network (GAN) as expounded by \cite{gan1} and \cite{gan2}. 
Recently, diffusion-based models \cite{diffusion,diffusion-old} have emerged as the epitome of excellence in image synthesis endeavors. 
DALLE-2 \cite{dalle2} employs a CLIP \cite{clip} text embedding to generate an image embedding via a prior network, which is then utilized by a diffusion decoder for image generation. 
Imagen \cite{imagen} turns to utilize T5-XXL \cite{t5xxl} to produce the text embedding.
Moreover, Stable Diffusion \cite{stablediffusion} trains diffusion models in latent space utilizing a pre-trained auto-encoder. 

The qualities of the images generated by these methods are greatly contingent upon the given text prompts. 
When users are not satisfied with the current results or wish to make modifications, our proposed \textit{DiffChat} serves as a powerful tool to facilitate interactive creation more easily.

\subsection{Instruction-Following Image Creation}
Traditional image editing models mainly targeted a single editing task such as style transfer \cite{tisold1,tisold2} or translation between image domains  \cite{tisold3,tisold4}. 
With the advent of CLIP and recent diffusion-based models, now users can guide image editing with text instructions \cite{instructpix2pix,couairon2023diffedit,zhang2023hive,InteractiveImageManipulation,BATINeT}.
\cite{InteractiveImageManipulation} focuses on the text-relevant content for manipulation and a super-resolution technique is applied. 
InstructPix2Pix \cite{instructpix2pix} releases an instruction-following image editing dataset and trains an end-to-end diffusion model. 
These studies mainly focus on the design of the image generation model structure, and thus \textit{DiffChat} can collaborate with the existing models for better user experience.
Lately, GILL \cite{gill}  and Emu \cite{emu} have introduced diffusion decoders into LLMs for end-to-end image generation. One issue with doing so is that whenever a better image decoder appears, the entire model needs to be re-trained and adapted, which often leads to unignorable costs.

Another alternative route 
is to design the forms and features of input texts \cite{prompt2prompt,wang2023instructedit,userfriendly,wei2023dialogpaint}. 
Prompt-to-Prompt \cite{prompt2prompt} manually designs several rules to control the text-to-image cross-attention for image editing. However, it is not automatic and cannot be directly performed using instructions.
InstructEdit \cite{wang2023instructedit} directly utilizes BLIP-2 \cite{blip2}  and ChatGPT \cite{chatgpt} to generate the original and target captions.  Yet, these texts still differ significantly from high-quality prompts used in real-world scenarios.
BeautifulPrompt \cite{DBLP:conf/emnlp/Cao0LWZ023} generates high-quality prompts by a language model but is not intended for image editing and conversations. 
PromptMagician~\cite{feng2023promptmagician} constructs a prompt recommendation model that identifies related prompt keywords for recommendations.
However, it is unable to explicitly follow user-defined instructions and the recommendations are also restricted by the database.
PRedItOR~\cite{ravi2023preditor} uses simplified target prompts to edit images but our \textit{DiffChat} directly uses user-defined instructions. 
DialogPaint \cite{wei2023dialogpaint} aims to train a language model to create instructions from conversations. Nevertheless, it is not as direct and effective as generating the target prompt by \textit{DiffChat}, which can be immediately used for image creation in collaboration with off-the-shelf TIS models such as the Stable Diffusion series.

\section{Methodology}
\begin{figure}[t]
\centering
\includegraphics[width=0.4\textwidth, height=0.38\textwidth]{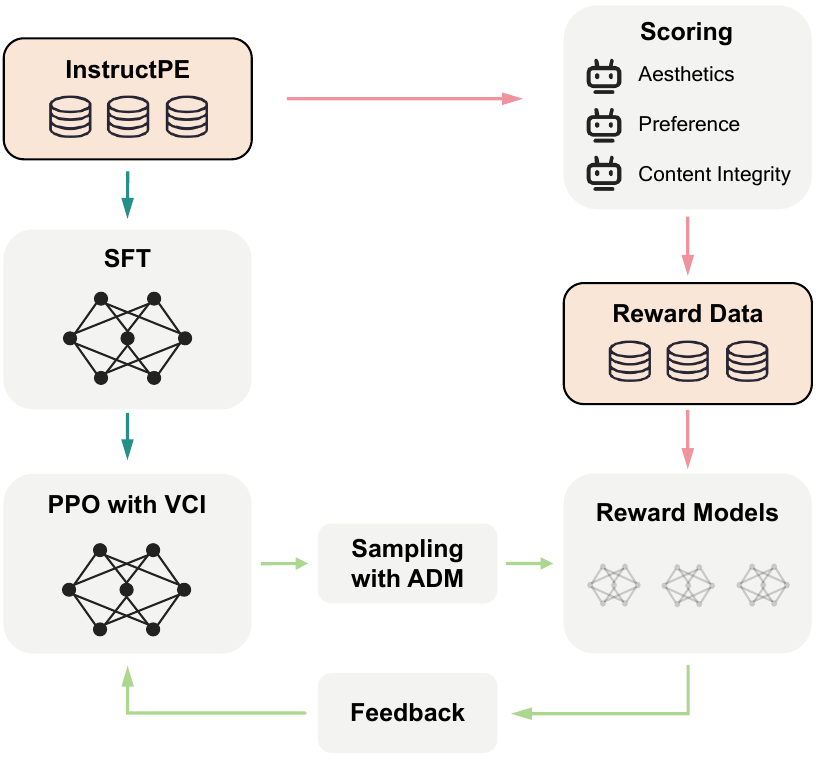}
\caption{The training procedure of~\textit{DiffChat}.} 
\label{method}
\end{figure}

The overall framework of our method is composed of three main steps: data collection of InstructPE  in Fig.~\ref{data}, supervised fine-tuning over \textit{DiffChat} and enhanced reinforcement learning with APC feedback in Fig.~\ref{method}.
We first construct the InstructPE  dataset from the raw data of InstructPix2Pix \cite{instructpix2pix} with prompt beautification and prompt engineering.
Next, \textit{DiffChat} is fine-tuned with supervised learning.
Finally, an enhanced PPO-based \cite{ppo} reinforcement learning process is performed with aesthetics, preference, and content integrity criteria.
The detailed explanations of each step are as follows.

\subsection{Data Collection of InstructPE}
The goal of the \textit{DiffChat} model is to generate the target prompts for interactive image creation given the raw prompts/images and the user-specified instruction.
To achieve it, we first need to build a highly-correlated dataset.
InstructPix2Pix \cite{instructpix2pix} has conducted a <\textit{raw prompt}, \textit{instruction}, \textit{target prompt}> format dataset using a fine-tuned GPT-3 \cite{gpt3} model. 
Specifically, it collects a relatively small dataset of editing triplets: input captions, edit instructions, output captions, and fine-tunes GPT-3 for the purpose. 
The input captions are sampled from the LAION-Aesthetics V2 6.5+ \cite{aesscore} dataset, and instructions and output captions are manually written. 
After this, a large number of new input captions are fed to the trained GPT-3 to generate instructions and output captions, resulting in the final 454,445 examples.
However, these input and output captions are simplified and user-friendly, which differ greatly from the effective and high-quality model-friendly prompts in practical applications with detailed descriptions and tags (examples can be found in Appendix \ref{promptexample}). 

\begin{figure}[t]
\centering
\includegraphics[width=0.48\textwidth,height=0.42\textwidth]{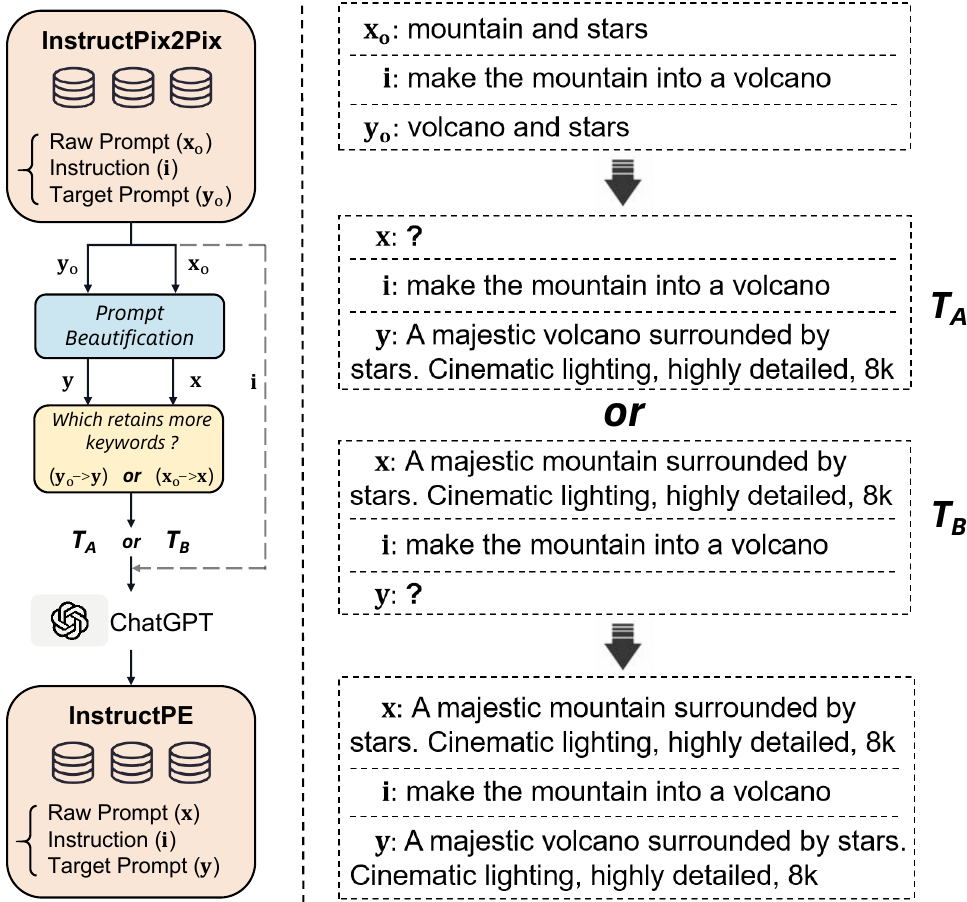}
\caption{Data collection process of~\textbf{InstructPE}.} 
\label{data}
\end{figure}

Realizing this, we first create a prompt beautification (PB) model to solve the problem.
We collect a large amount of real-world high-quality prompts from DiffusionDB\footnote{https://huggingface.co/datasets/poloclub/diffusiondb}, MagicPrompt\footnote{https://huggingface.co/datasets/Gustavosta/Stable-Diffusion-Prompts}, and the Civitai website\footnote{https://civitai.com}.
Next, we ask ChatGPT to summarize these high-quality prompts into simplified user-friendly prompts\footnote{The detailed prompts and examples to ask ChatGPT for prompt beautification can be found in Appendix \ref{pb}.} (such as the captions in the InstructPix2Pix data as shown in Fig.~\ref{data}). 
Through this approach, we have obtained a plentiful supply of <\textit{simplified}, \textit{high-quality}> prompt pairs, which will be used to fine-tune a BLOOM-1.1B \cite{bloom} model as our PB model.

After obtaining the PB model, the collection process of InstructPE  is allowed to begin, as shown in Fig.~\ref{data}. 
The raw prompt $\mathbf{x_o}$ and target prompt $\mathbf{y_o}$ of InstructPix2Pix are separately sent to our PB model to generate the beautified $\mathbf{x}$ and $\mathbf{y}$.
Next, due to the inevitable risk of missing keywords during the generation process of the PB model, 
we decide whether to use template \textbf{$T_A$} or \textbf{$T_B$} for the next step of interaction with ChatGPT based on which group $\mathbf{(y_o}$ --> $\mathbf{y})$  or $\mathbf{(x_o}$ --> $\mathbf{x})$  retains more keywords (the keywords extraction process is shown as the function in Line 10 of Lst.~\ref{ci} in Appendix \ref{app_ci}).
The reason behind this operation is that we hope to maintain consistency between the modified prompts and the original prompts as much as possible.
For example, if $\mathbf{(y_o}$ --> $\mathbf{y})$ retains more keywords than $\mathbf{(x_o}$ --> $\mathbf{x})$, we will set $\mathbf{y}$ as the known reference and let ChatGPT generate $\mathbf{x}$, and vice versa.
Then, given $\mathbf{x}$ or $\mathbf{y}$ and the instruction $\mathbf{i}$, we ask ChatGPT to write another one ($\mathbf{y}$ or $\mathbf{x}$)
with prompt engineering\footnote{The detailed prompts and examples to ask ChatGPT for InstructPE can be found in Appendix \ref{ask}.}.
Finally, our  InstructPE dataset is organized as ($\mathbf{x}$, $\mathbf{i}$, $\mathbf{y}$).

Moreover, post-processing is also involved. Non-English and NSFW (not safe for work)  examples are first filtered out. 
Next, we utilize the scoring models that will be discussed in Sec.~\ref{rewardmodel} to filter out the low-quality examples. 
We finally collect 234,786 triplets as our training set and 5,582 triplets as the testing set.

\subsection{Supervised Fine-Tuning}
Given the InstructPE  dataset with triplets $D=\{(\mathbf{x},\mathbf{i},\mathbf{y})\}$ containing input prompts $\mathbf{x}$, instructions $\mathbf{i}$,  and target prompts $\mathbf{y}$, we fine-tune a decoder-only language model to output each high-quality prompt of tokens $\mathbf{y} = \{y_1, ..., y_t \}$, where $t$ is the length of $\mathbf{y}$.
We use the auto-regressive language modeling objective to maximize the following likelihood~\cite{gpt2}:

\begin{small}
\begin{align}
\mathcal{L}^{SFT} = - \sum_t \log P(y_t\mid \mathcal{T}(\mathbf{x}, \mathbf{i}), y_1, ..., y_{t-1}),
\end{align}
\end{small}
where $\mathcal{T}$ is a template for organizing $\mathbf{x}$ and  $\mathbf{i}$ into a prefix sentence\footnote{We set the template $\mathcal{T}$ as  ``\texttt{Instruction: Give a description of the image and a modification to generate a drawing prompt.\textbackslash nInput: \{$\mathbf{x}$\}\textbackslash nModification: \{$\mathbf{i}$\}\textbackslash nOutput:}''.}.

\subsection{Reinforcement Learning with Feedback}
As the collected dataset inevitably contains noises, e.g., the target prompts do not strictly follow the corresponding input prompts and instructions, 
the performance of the supervising trained model can be unsatisfactory. 
To make further development, we aim to follow \cite{instructgpt} to perform the task using reinforcement learning leveraging the proximal policy optimization (PPO) \cite{ppo} algorithm.
Yet before that, we propose the following improvements to adapt it to our task.

\paragraph{Reward Models}\label{rewardmodel}
The agent model needs to obtain reward feedback from the environment to update its policy in the desired direction.
Focusing on our tasks, rewards must reflect the aspects that users care about the results of interactive image creation.
In this regard, we design three user-concerned criteria:
(1) \textit{Aesthetics}. It represents the aesthetic evaluation of the created images.
(2) \textit{Preference}. It indicates the user's preference for the specified image relative to other ones.
(3) \textit{Content integrity}. It evaluates the completeness of the key contents of input prompts and instructions contained in the creations.
However, even though using human feedback to meet these standards may often bring promising results \cite{instructgpt}, it requires extensive and tedious labor efforts. 
Instead, \citet{cai} proposes to use AI models to instruct the training of LLMs.
Inspired by this, we also aim to use off-the-shelf AI models along with self-designed heuristic rules to automatically score our generated results, thus avoiding the cost of expensive human labeling.
Specifically, the aesthetic score \cite{aesscore} and PickScore~\cite{pickscore} are considered as our \textit{aesthetics} 
 and \textit{preference}  criteria, respectively. 
Moreover, we design the \textit{content integrity} score:
Given ($\mathbf{x_o},\mathbf{i},\mathbf{y_o}$) as references, 
it first heuristically extracts keywords from $\mathbf{y_o}$ and identifies the highlighted ones. 
Then it determines whether to reward based on whether the content integrity of $\mathbf{y}$ reaches a threshold.
A code example in Python style is shown in Appendix \ref{app_ci}.

\paragraph{Action-space Dynamic Modification (ADM)}
The action spaces involved in language generation often far surpass the capabilities of most discrete action spaces in traditional designs \cite{mnih2015human,hessel2018rainbow}. For instance, GPT-3 \cite{gpt3} and T5 \cite{t5xxl} models have vocabulary sizes of 50K and 32K respectively. 
During vanilla off-policy data sampling $\mathrm{\pi^o}$, it randomly selects the next token $y_t$ based on the probability distribution over the entire action/vocabulary space $\mathcal{Y}$:
\begin{align}
y_t \sim \mathrm{\pi^o} (\cdot \mid P (\mathcal{Y} \mid \mathcal{T}(\mathbf{x}, \mathbf{i}), y_1, ..., y_{t-1})).
\end{align}
In this regard, every token with a non-zero probability has a chance of being selected.
However, the large size of the action space is a fundamental reason for the instability of sample qualities. 
To tackle this, we introduce action-space dynamic modification (ADM) to simultaneously refine both positive and negative samples.
For positive ones, we exclude 
tokens with less information quantity from the action space $\mathcal{Y}$ to form $\mathcal{\widehat{Y}}^+_t$ during each sampling step $t$:
\begin{align}
y_t \sim \pi^+ (\cdot \mid P (\mathcal{\widehat{Y}}^+_t \mid \mathcal{T}(\mathbf{x}, \mathbf{i}), y_1, ..., y_{t-1})),
\end{align}
where achieving $\mathcal{\widehat{Y}}^+_t$  involves employing 
locally typical sampling~\cite{typicalp} with probability $p$, 
which restricts tokens to the smallest set while ensuring the sum of their probabilities surpasses the specified probability parameter $p$. 
For negative samples, we conduct that:
\begin{align}
y_t \sim \pi^- (\cdot \mid P  (\mathcal{\widehat{Y}}^-_t \mid \mathcal{T}(\mathbf{x}, \mathbf{i}), y_1, ..., y_{t-1})),
\end{align}
where we randomly select keywords (from the results of the function in Line 10 of Lst.~\ref{ci} in Appendix \ref{app_ci}) for each of a small proportion of (with a 4\% probability)
target prompts and then remove (with a 50\% probability) or modify (with a 50\% probability) it 
to create $\mathcal{\widehat{Y}}^-_t$.
For the modification, given the original keyword, we select a keyword with the same part of speech but not in its synonym dictionary (Line 36 of Lst.~\ref{ci}) in the training set.
In this way, we can simulate the real omission or replacement errors to enable targeted optimization.

\begin{table*}[htb]
\centering
\begin{small}
\setlength{\tabcolsep}{1.8mm}{
\begin{tabular}{lccccccc}
\toprule
    \textbf{Method} &\textbf{PickScore $\uparrow$} & \textbf{Aes. Score $\uparrow$} & \textbf{HPS $\uparrow$}& \textbf{ CLIP-S $\uparrow$} & \textbf{ D-CLIP-S $\uparrow$}& \textbf{CI Score $\uparrow$}  & \textbf{Avg. Rank. $\downarrow$}\\
\hline
ChatGPT & 19.569
& \underline{6.394}
& 20.822
& \underline{29.354}
& 15.802
& \textbf{87.496}
& \underline{2.500} \\
\hline
InstructPix2Pix&  19.346
& 5.726
& 19.036
& 22.670
& \textbf{17.327} 
& - & 3.400 \\
\hline
\emph{DiffChat} (SFT only) & \underline{19.571}
& 6.392
& \underline{20.836}
& 29.350
& 16.596
& 85.089
& 2.667  \\
\emph{DiffChat} (full imp.)  & \textbf{19.584}
& \textbf{6.416}
& \textbf{20.851}
& \textbf{29.397}
& \underline{16.787}
& \underline{87.314}
& \textbf{1.333}  \\
\bottomrule
\end{tabular}}
\end{small}
\caption{Average automatic evaluation results on the InstructPE testing set with different SD models. 
 More detailed results are shown in Appendix~\ref{append_detail_result}. 
Avg. Rank. is calculated as the average ranking value under each score. Aes. Score: the aesthetic score.
CLIP-S: CLIP score.
D-CLIP-S: directional CLIP similarity. 
SFT only: only conducting supervised fine-tuning.
Full imp.: Full implementation.
}
\label{result}
\end{table*}

\begin{figure*}[t]
\centering
\includegraphics[width=0.98\textwidth]{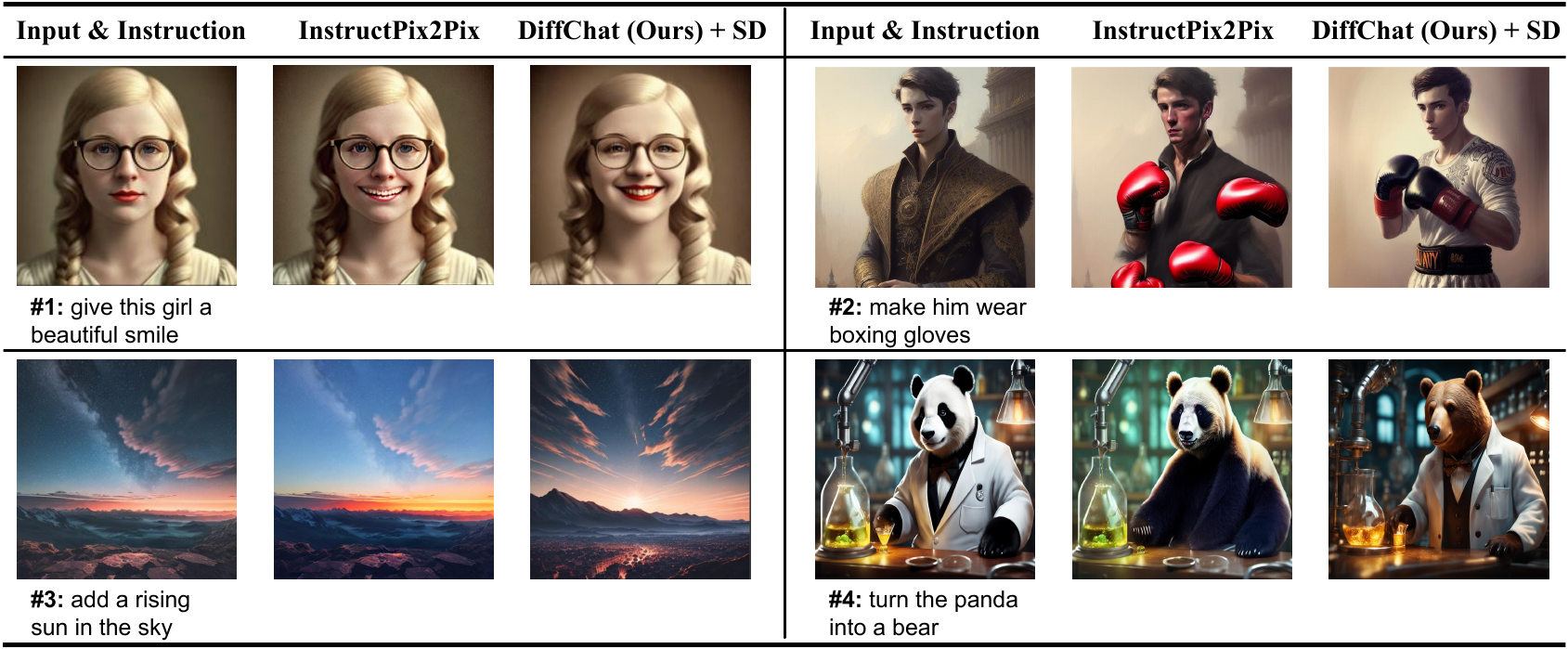}
\caption{Qualitative results of InstructPix2Pix and \textit{DiffChat} + SD for instruction-following image creation. 
}
\label{visres}
\end{figure*}

\paragraph{Value Estimation with Content Integrity (VCI)}
Advanced policy gradient methods \cite{trpo,ppo} introduce the advantage function $A$ to measure the extent to which an action $a_t$ is better or worse than the policy’s average action in a particular state $s_t$. Generalized advantage estimation (GAE) \cite{gae} is widely adopted to calculate $A$ as:
\begin{align}
A(s_t,a_t) &= \sum_l (\gamma \lambda)^l\delta_{t+l},\\
\mathrm{where} \quad \delta_t &= r_t+\gamma V(s_{t+1}) - V(s_t).
\end{align}
Here, $l$ is the trajectory length, $\lambda$ and $\gamma$ are trade-off and discount parameters. $r_t$ is the reward in $t$-th step. $V$ is the value function to comprehensively evaluate the current state. 
In order to help it better perceive the current text generation progress, we propose to add content integrity to compute the value for reinforcement learning:
\begin{align}
\widehat{V}(s_t)=V(s_t)+\alpha\cdot\mathrm{CI\_score}(s_t).
\end{align}
$\alpha$ is a trade-off hyper-parameter.
$\mathrm{CI\_score}$ has been introduced and defined in Lst.~\ref{ci}, 
and $s_t$ is formulated as ($\mathbf{x_o},\mathbf{i},\mathbf{y_o}, \mathbf{y}_{\sim t}=\{y_1,...,y_t$\}).

\section{Experiments}
\subsection{Implementation Details}
\paragraph{Training Settings} We use the pre-trained checkpoint of BLOOM \cite{bloom} 1.1B parameters with 24 transformer layers as the backbone of~\textit{DiffChat}. 
Note that, choosing this relatively small version is to ensure the high inference efficiency to support real-world applications. 
Our method is independent of the selection of specific models and we find
that the 1.1B model is sufficient to achieve satisfactory performance.
The BFLOAT16 format is leveraged to save GPU memory and speed up training. 
All the experiments are implemented in PyTorch and run on a single server with NVIDIA Tesla A100 GPUs. 
More detailed parameters are shown in Appendix \ref{param}.

\begin{table*}[t]
\centering
\scalebox{0.9}{
\begin{tabular}{p{0.02cm}p{4cm}p{2cm}p{4.5cm}p{4.5cm}}
\toprule
\small{\textbf{\#}} & \small{\textbf{Raw Prompt}} & \small{\textbf{Instruction}} & \small{\textbf{DiffChat w/o. $\pi^-$}} & \small{\textbf{DiffChat}}  \\
\hline
1& \small{A \textcolor{gray}{digital painting} of a landscape of
..., highly detailed}
 & \small{make it a painting}
 & \small{A \textcolor{gray}{digital painting} of a landscape of 
 ..., highly detailed}
 & \small{A \textcolor{gray}{painting} of a landscape of 
 ..., highly detailed}
\\
\hline
2&  \small{Photograph of a wedding. ... prewedding \textcolor{gray}{photorealistic}}
 & \small{have a vintage feel}
 & \small{Photograph of a \textcolor{gray}{vintage} wedding. ... prewedding, \textcolor{gray}{photorealistic}}
 & \small{Photograph of a \textcolor{gray}{vintage} wedding. ... prewedding, \textcolor{gray}{with a vintage feel}}
\\
\bottomrule
\end{tabular}}
\\(a) Ablation study for ADM $\pi^-$.
\bigskip 

\scalebox{0.9}{
\begin{tabular}{p{0.02cm}p{4cm}p{2cm}p{4.5cm}p{4.5cm}}
\toprule
\small{\textbf{\#}} &\small{\textbf{Raw Prompt}} & \small{\textbf{Instruction}} & \small{\textbf{DiffChat w/o. VCI}} & \small{\textbf{DiffChat}}  \\
\hline
1& \small{study of fair hair \textcolor{gray}{beauty} ..., full hd } & \small{turn the woman into a cat }
 & \small{study of fair hair \textcolor{gray}{beauty} ..., full hd, \textcolor{gray}{cat}}
 & \small{study of fair hair \textcolor{gray}{cat} ..., full hd}
\\
\hline
2& \small{A wildflower ... in washington D. c, by Greg Rutkowski}
 & \small{have it be in the snow}
 & \small{A wildflower ... in washington D. c, by Greg Rutkowski, \textcolor{gray}{in the snow}}
 & \small{A wildflower ... in washington D. c, \textcolor{gray}{covered in snow,} by Greg Rutkowski }\\
\bottomrule
\end{tabular}}
\\(b) Ablation study for VCI.
\caption{Ablation study results for ADM $\pi^-$ and VCI.}
\label{ablation}
\end{table*}

\paragraph{Evaluation Protocols} Systematically evaluating the goodness of a prompt engineering model is a challenging task. One of the most straightforward methods is to evaluate the images generated by the prompts that models produce.
We use Stable Diffusion 1.5\footnote{https://huggingface.co/runwayml/stable-diffusion-v1-5},
Deliberate\footnote{https://huggingface.co/XpucT/Deliberate}, Dreamlike\footnote{https://huggingface.co/dreamlike-art/dreamlike-photoreal-2.0},  Realistic\footnote{https://huggingface.co/SG161222/Realistic\_Vision\_V1.4},
and Stable Diffusion XL 1.0\footnote{https://huggingface.co/stabilityai/stable-diffusion-xl-base-1.0}
with fixed seeds to generate images and calculate PickScore \cite{pickscore}, the aesthetic score \cite{aesscore}, HPS \cite{hps},
CLIP score \cite{clipscore}, 
directional CLIP similarity \cite{directclip} and our CI Score before thresholding for the images and the corresponding prompts. 
Furthermore, we also conduct human evaluations on 100 randomly selected examples from the testing set and 100 randomly user-written examples. 
Given the raw images and instructions, we ask human experts to pick the most desirable target images\footnote{The user interface is shown in Appendix \ref{interface}.} generated by the different methods and report the win rates of \textit{DiffChat}.

\begin{figure}[t]
\centering
\includegraphics[width=0.45\textwidth, height=4.cm]{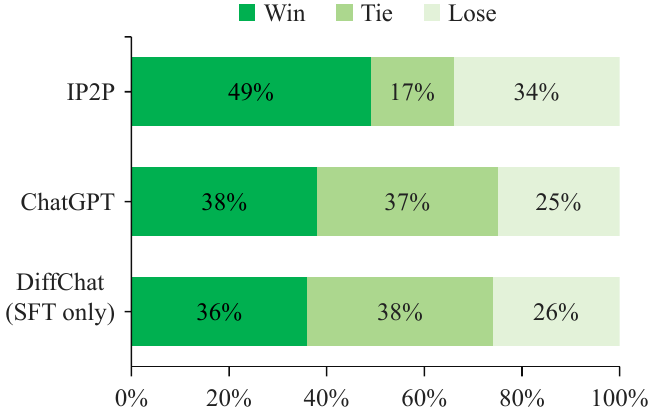}
\caption{Results of human preference evaluation (i.e.,
Win/Tie/Lose rates of our method against others). IP2P is short for InstructPix2Pix.} 
\label{humnaeva}
\end{figure}

\subsection{Overall Performance}
\paragraph{Competitors}
We consider two strong competitors: ChatGPT \cite{chatgpt} and InstructPix2Pix \cite{instructpix2pix}. 
ChatGPT is almost the most powerful general-purpose LLM with astonishing few-shot learning abilities.
Here, it serves as a prompt modifier given the raw prompt and instruction.
InstructPix2Pix is a popular end-to-end instruction-following image editing model trained from self-collected data.

\begin{figure*}[t]
\centering
\includegraphics[width=0.9\textwidth, height=8.7cm]{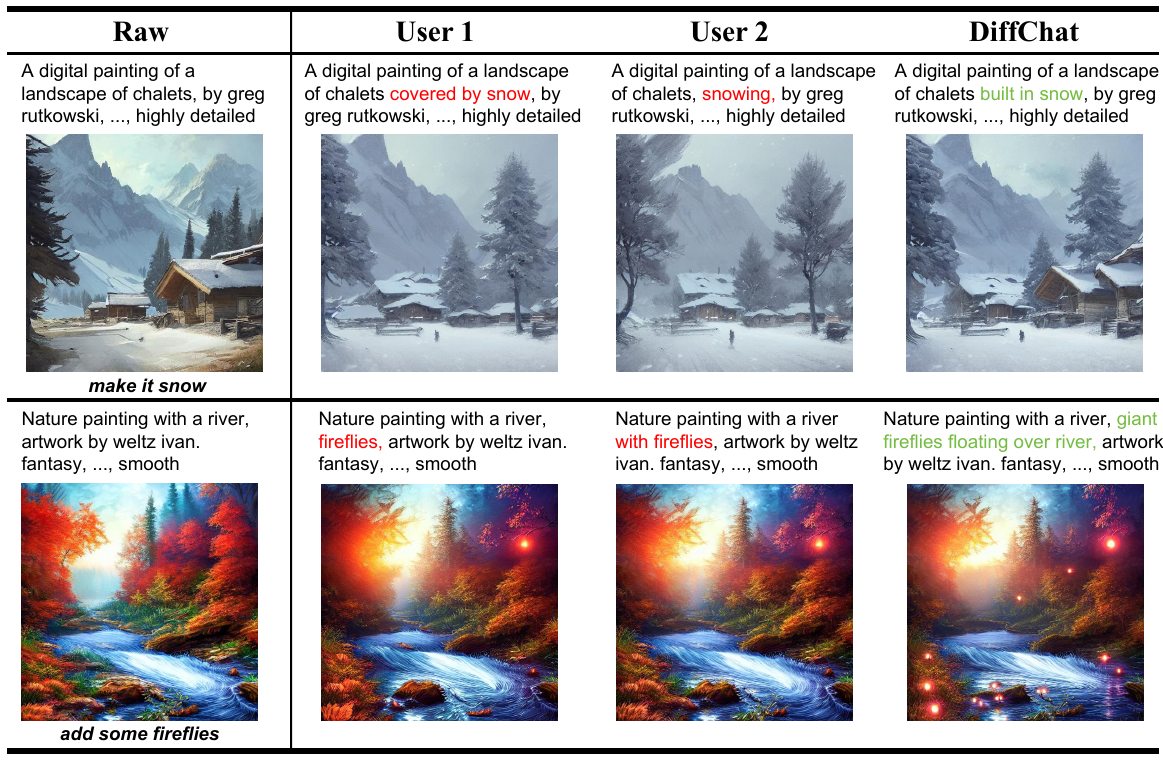}
\caption{Comparison between \textit{DiffChat} and human prompt writers given the same inputs and instructions.} 
\label{vsuser}
\end{figure*}

\begin{figure}[t]
\centering
\includegraphics[width=0.45\textwidth]{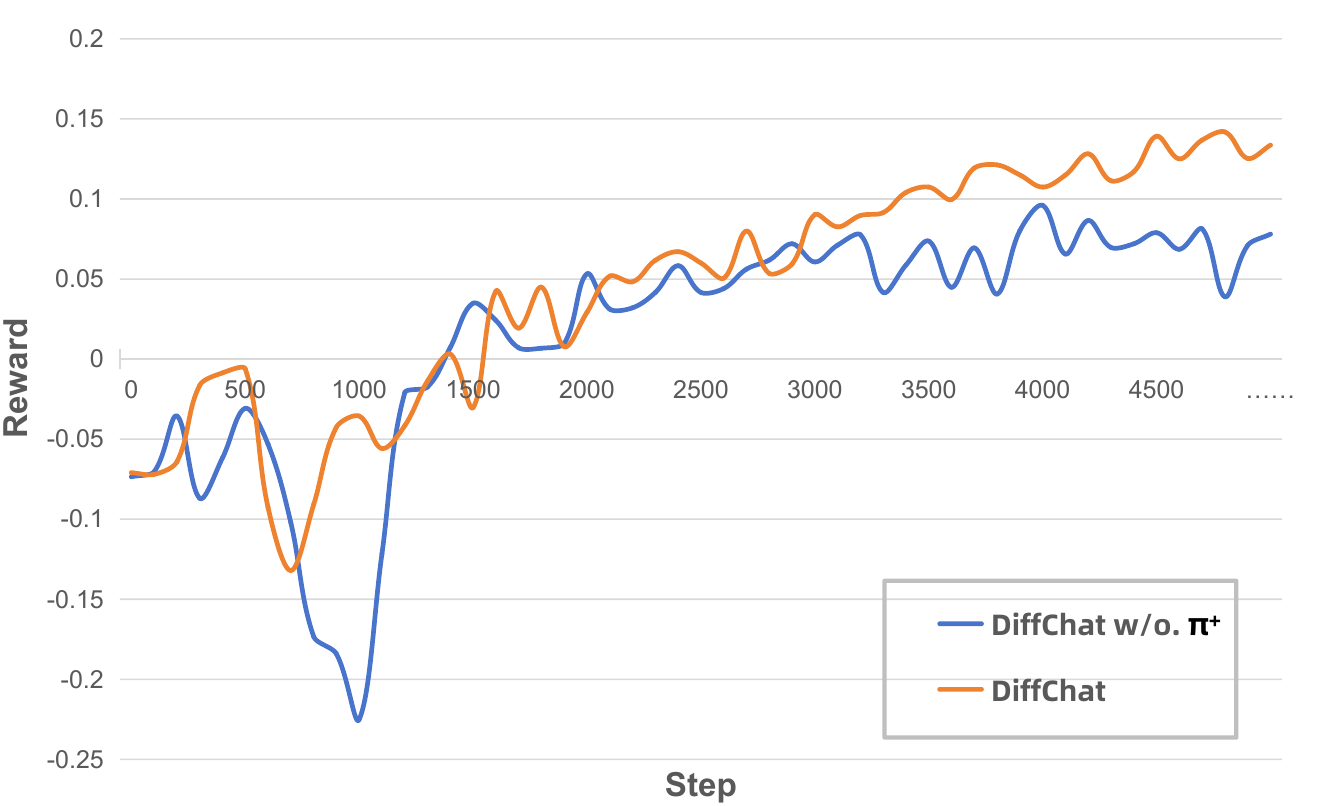}
\caption{Ablation study for ADM $\pi^+$.} 
\label{rewardabl}
\end{figure}

From Tab.~\ref{result} with the average automatic evaluation results on different SD models, our method consistently achieves competitive or superior performances in most scores.
As D-CLIP-S reflects how much the change in text prompts agrees with the change in the images, InstructPix2Pix which directly edits the concerned parts and maintains the overall structure of the image naturally reaches the highest score.
We can have an ahead look at example \#2 in Fig.~\ref{visres}: InstructPix2Pix mainly adds boxing gloves in the lower local area with an unnatural presentation. 
On the contrary, our method can add boxing gloves and change it to a boxing posture to fully demonstrate ``wear boxing gloves''. 
However, in this case, its D-CLIP-S is still lower than InstructPix2Pix's.
CI score mainly reflects the completeness of keywords contained in the target prompt. Although ChatGPT can achieve a relatively high score, it leads to even less beautiful creations.
More analyses are shown in Appendix~\ref{append_detail_result}.

As shown in  Fig.~\ref{humnaeva}, the human evaluation experiment also indicates the superiority of our approach. \textit{DiffChat} has the highest positive recognition rate among evaluators compared with other models.

Fig.~\ref{visres} presents the qualitative results generated by InstructPix2Pix and \textit{DiffChat} + SD. 
For example, given the instruction ``give this girl a beautiful smile'' in \#1, InstructPix2Pix mainly focuses on modifying local areas of the mouth, while ignoring other parts of the face. On the contrary, \textit{DiffChat} + SD can achieve muscle modifications in the eyes, cheeks, and mouth areas, resulting in more natural and beautiful image creations.
In \#3, InstructPix2Pix mainly increases the overall brightness of the image to express ``a rising sun'', yet \textit{DiffChat} + SD achieves the addition of half a dazzling sun on the horizon to reflect the rising process.
From these examples, we can find that \textit{DiffChat} + SD can complete more aesthetically pleasing creations based on instructions with only minor out-of-concern details changing.
Compared with models that directly edit images such as InstructPix2Pix, it can avoid the collapse of local areas in the creation.

\subsection{Detailed Analysis}
\paragraph{Ablation Study.}
From Fig.~\ref{rewardabl} we can find that ADM $\pi^+$ improves the training stability during optimization and achieves steady higher reward feedback.
Furthermore, Tab.~\ref{ablation} (a) shows the improvement brought by ADM $\pi^-$.
It helps the model better train with hard negative samples and try to avoid issues such as incorrect replacement and omissions in real applications.
From Tab.~\ref{ablation} (b), we can also infer that the vanilla model tends to simply insert the content that needs to be replaced or added at the end of the raw prompts. With the help of VCI, \textit{DiffChat} can be corrected to prefer generating key information content earlier to alleviate it.

\paragraph{Does \textit{DiffChat} Perform Better than Human Prompt Writers?}
We further explore whether \textit{DiffChat} can more effectively bring better image creation experience than users themselves who write prompts completely.
For example in Fig.~\ref{vsuser} \#2, given the raw image of a river, if the users want to \textit{add some fireflies}, they may modify the prompt as ``..., fireflies, ...'' or ``..., with fireflies, ...''. However, the resulting images only add a firefly in the top right corner. On the contrary, the modification made by \textit{DiffChat} can lead to a better image creation which is more in line with user expectations.

\paragraph{Is \textit{DiffChat} Transferable across Different TIS Models?}
To verify the transferability of \textit{DiffChat} across different TIS models, we also consider other diffusion-style popular models such as  Deliberate, Dreamlike, and Realistic (see them in Appendix~\ref{multiple-versions}). 
Since our pipeline utilizes prompts as intermediaries for interactive image creation, its flexibility and generalization are guaranteed.
More qualitative examples are shown in Appendix \ref{multiple-versions}.

\section{Conclusion}
In this paper, we propose \textit{DiffChat} to follow user-specified instructions to interact with TIS models for image creation.
We collect and release the~\textbf{InstructPE} dataset for training instruction-following prompt engineering models. 
A reinforcement learning framework with aesthetics, preference, and content integrity feedback is introduced to align supervising fine-tuned LLMs.
Action-space dynamic masking and value estimation with content integrity are also involved for further improvement.
Extensive experimental results show that~\emph{DiffChat} outperforms competitors in terms of both automatic and human evaluation.

\section*{Limitations}
Although~\emph{DiffChat} can produce prompts of aesthetically pleasing images given instructions, limited by the training data, it has the risk of ignoring minor parts of the information in the original prompts. 
Furthermore, since~\emph{DiffChat} is guided by the APC feedback as introduced in Sec.\ref{rewardmodel} during reinforcement learning, the choices of specific implementation approaches will affect the upper bound of the model performance.
These improvements are left to our subsequent work.

\section*{Ethical Considerations}
The techniques for training the~\emph{DiffChat} model presented in this work are fully methodological, thereby there are no direct negative social impacts of our method.
Additionally, we have filtered out NSFW prompts from our training data to ensure that the generated contents are suitable for public distribution. 
However, given the inherent challenges in controlling the generative process, there is a slight possibility (though improbable) for our model to produce toxic contents. 
We advise users to refrain from utilizing~\emph{DiffChat} intentionally to generate offensive or inappropriate images, and emphasize the need for responsible consideration of potential risks for online deployment.

\section*{Acknowledgements}

This work is partially supported by
Alibaba Cloud Group, through Research Talent Program with South China University of Technology.

\bibliography{anthology,custom}

\appendix

\section{Appendix}

\subsection{Examples of User-Friendly and Model-Friendly Prompts}\label{promptexample}
\begin{figure}[h]
\centering
\includegraphics[width=0.49\textwidth]{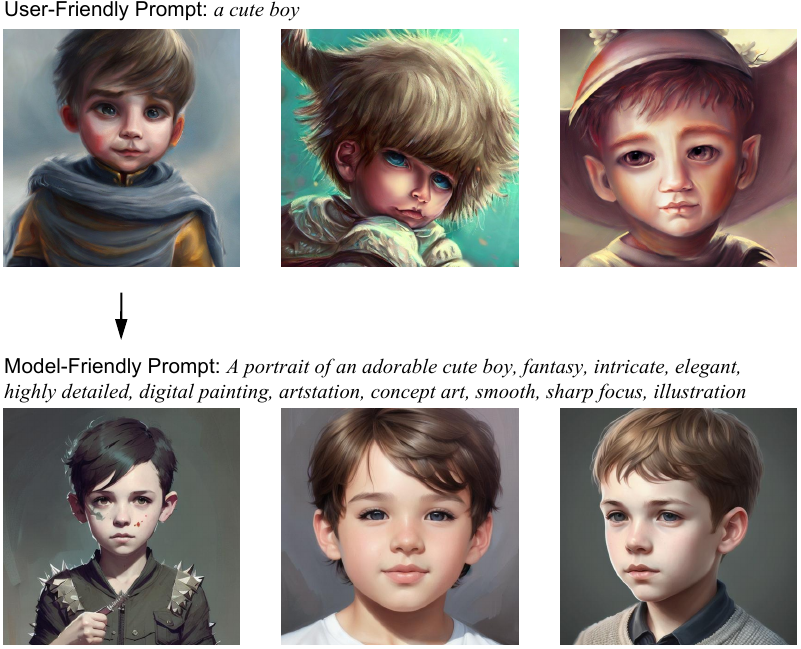}
\caption{Example of user-friendly 
 and model-friendly prompts.} 
\label{pea}
\end{figure}

As shown in Fig.~\ref{pea}, the user-friendly prompt is usually simplified and short, while the model-friendly prompt in practical usage is detailed and long with several descriptions and tags. They can lead to image creations with completely different qualities.

\subsection{Asking ChatGPT for Prompt Beautification}\label{pb}
\begin{figure}[h]
\centering
\includegraphics[width=0.49\textwidth]{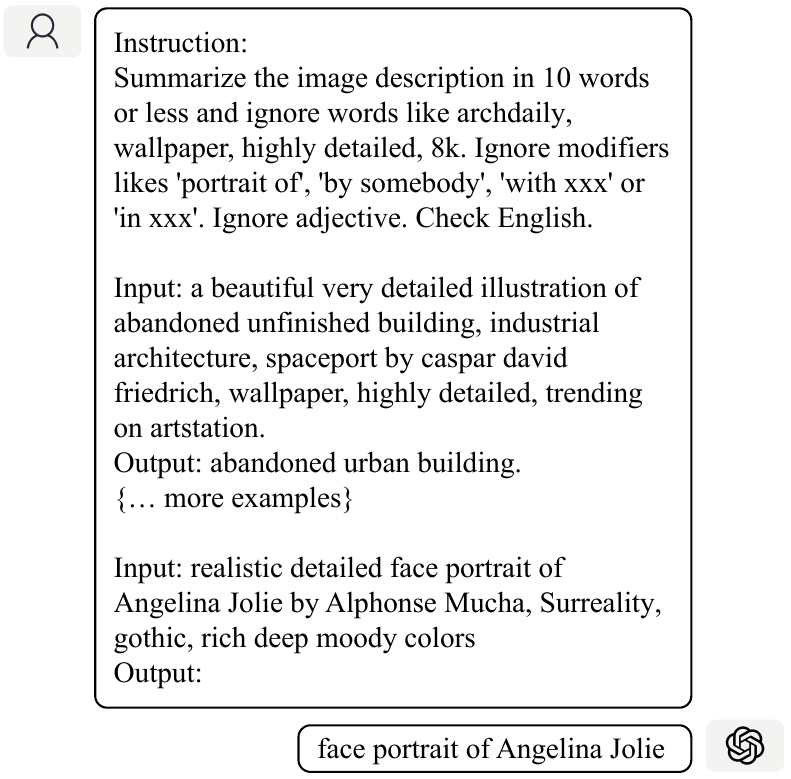}
\caption{Asking ChatGPT for generating prompt beautification data.} 
\label{append1}
\end{figure}

Fig.~\ref{append1} shows an example of asking ChatGPT to generate data for prompt beautification.

\subsection{Asking ChatGPT for InstructPE}\label{ask}
\begin{figure}[h]
\centering
\includegraphics[width=0.4\textwidth]{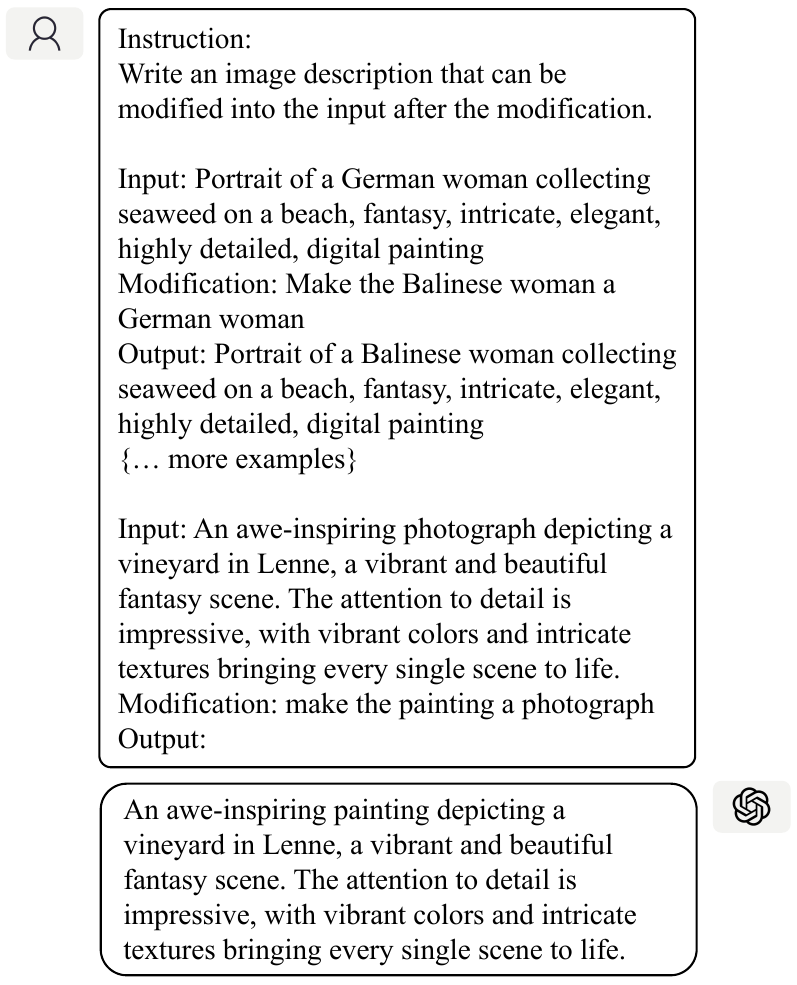}
\\(a) Given $\mathbf{y}$ and $\mathbf{i}$ to generate $\mathbf{x}$.\\
\bigskip
\includegraphics[width=0.4\textwidth]{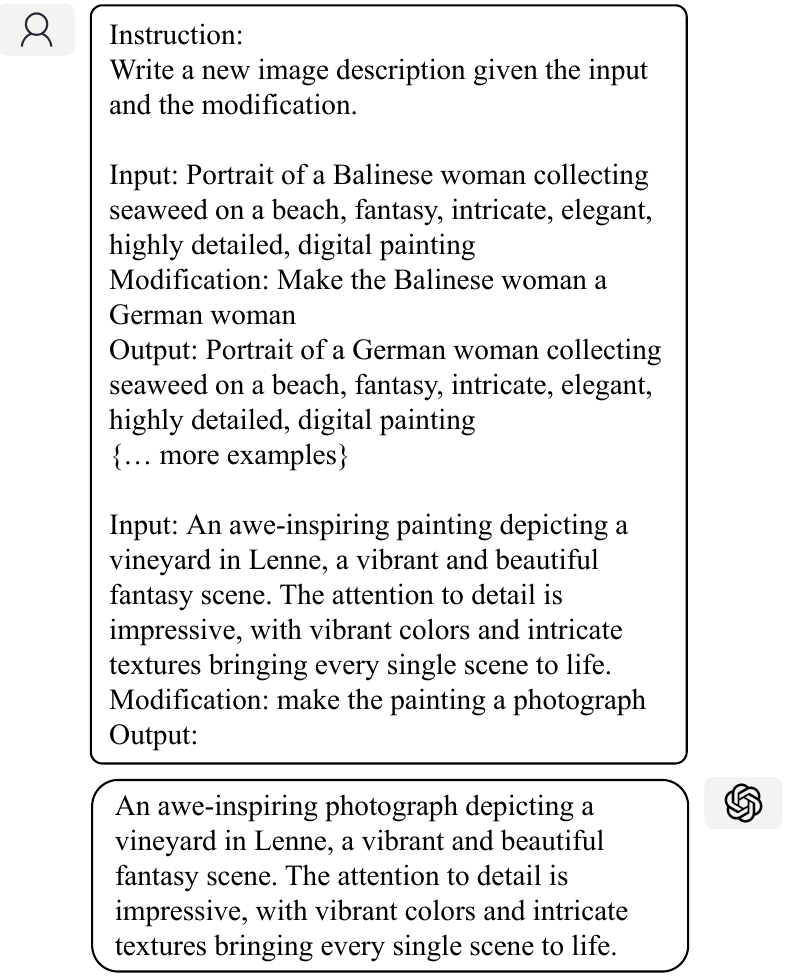}
\\(b) Given $\mathbf{x}$ and $\mathbf{i}$ to generate $\mathbf{y}$.
\caption{Asking ChatGPT for generating~\textbf{InstructPE} data.} 
\label{append2}
\end{figure}

Fig.~\ref{append2} shows an example of asking ChatGPT to generate data for~\textbf{InstructPE}.

\subsection{Content Integrity Calculation}\label{app_ci}
A code example in Python style for the calculation of content integrity score is shown in Lst.~\ref{ci}. 
Some details are abbreviated for better readability.

\begin{lstlisting}[language=Python, caption=Pseudocode of content integrity score calculation., label=ci]
# Pseudocode of CI Score Calculation
import nltk

# Word lemmatization.
def lemmatize(words):
    # "wnl" is a word lemmatization model referenced from https://www.nltk.org/_modules/nltk/stem/wordnet.html.
    return [wnl.lemmatize(w[0], w[1]) for w in words]

# Keywords extraction.
def extract_keywords(text):
    words = nltk.pos_tag(split_to_words(text))
    # "candidates" contains parts of speech that are identified as keywords.
    words = [word for word in words if word[1] in candidates]
    return list(set(lemmatize(words)))

# Input:
# x_o: Raw Prompt of InstructPix2Pix.
#   i: Instruction of InstructPix2Pix.
# y_o: Target Prompt of InstructPix2Pix.
#   y: Target Prompt of InstructPE.
def CI_score(x_o, i, y_o, y, thres=0.7):
    # Split prompt into words and conduct POS tagging.
    y_words = nltk.pos_tag(nltk.split_to_words(y))
    # Word lemmatization.
    y_words = lemmatize(y_words)
    # Extract keywords of x_o, i, and y_o.
    x_o_keywords = extract_keywords(x_o)
    i_keywords   = extract_keywords(i)
    y_o_keywords = extract_keywords(y_o)
    # Identify highlighted words.
    highlighted = [q for q in y_o_keywords if q not in x_o_keywords and q in i_keywords]
    # Start calculation.
    cnt = 0
    for y in y_o_keywords:
      # Retrieve in synonym dictionary.
      syns = synonym_dict(y)
      for s in syns:
        if s in y_words:
          if s in highlighted:
            # Give highlighted words greater weight.
              cnt += 2
          else:
              cnt += 1
          break
    CI_score = min(cnt/len(y_o_keywords), 1.)

    # Thresholding.
    return CI_score if CI_score >= thres else 0

\end{lstlisting}

\subsection{More Training Details  }\label{param}
Most of the detailed training parameters are listed in Tab.~\ref{trainparam}.
Moreover, for the reinforcement learning, 
we set the initial KL coefficient as 0.05, 
$\gamma$  as 0.99, $\lambda$ as 0.95, $p$ as 0.97, and $\alpha$ as 0.05, respectively. 
We use the AdamW optimizer wth eps 1e-8 and ($\beta_1$, $\beta_2$) = (0.9, 0.95).
Cosine annealing schedule is adopted.
Clipping range for PPO policy loss is set as 0.2.
 Value loss scale w.r.t policy loss is set as 0.5.

\begin{table}[h]
\centering
\begin{small}
\begin{tabular}{lccc}
\toprule
\textbf{Parameters} &\textbf{SFT} & \textbf{RMs} & \textbf{RL} \\
\hline
Epoch & 3 & 1 & 5 \\
Batch Size & 64 & 64 & 128 \\
Maximum Length & 384 & 384 & 384 \\
Learning Rate & 2e-5 & 5e-6 & 5e-6 \\
Unfreezing Layers & All & All & Last 8 layers \\
Weight Decay & 0 & 1e-3 & 1e-6 \\

\bottomrule
\end{tabular}
\end{small}
\caption{Detailed training parameters for SFT (supervised fine-tuning), RMs (reward models), and RL (reinforcement learning) parts.}
\label{trainparam}
\end{table}

\subsection{Interface for Human Preference Evaluation}\label{interface}
\begin{figure}[h]
\centering
\includegraphics[width=0.45\textwidth, height=8.5cm]{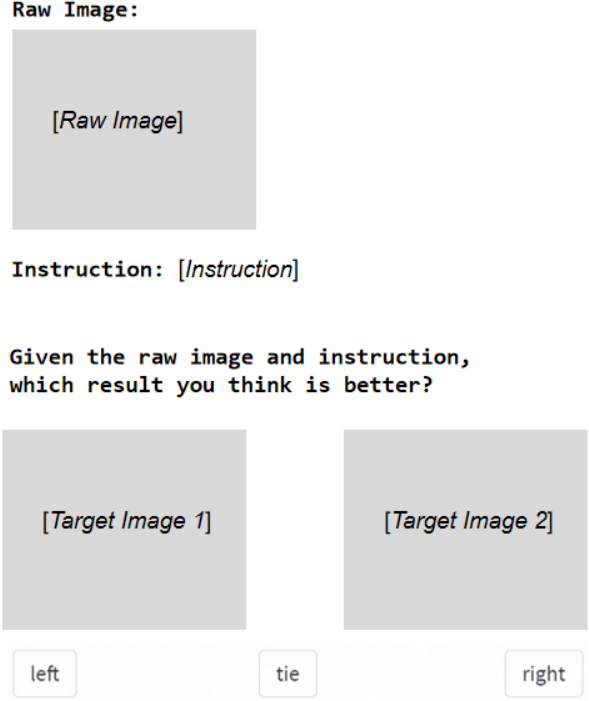}
\caption{Interface for the human evaluation.} 
\label{pickimg}
\end{figure}

Fig.~\ref{pickimg} shows a screenshot of the interface for the human preference evaluation experiment.

\subsection{Detailed Automatic Evaluation Results}\label{append_detail_result}
In this subsection, we provide the detailed automatic evaluation results as shown in Tab.~\ref{detail_result} utilizing different SD models, such as Stable Diffusion 1.5, Deliberate, Dreamlike, Realistic, and Stable Diffusion XL 1.0.
It can be found that our method achieves the highest average ranking when collaborating with multiple different SD models.
Another noteworthy fact is that InstructPix2Pix is optimized for the D-CLIP-S metric based on Stable Diffusion 1.5 model. 
When evaluated under other metrics, it performs relatively poorly, 
and the application on the new Stable Diffusion XL 1.0 leads to significantly poorer results, which also reflects the limitations of its technical route's generalizability.

\begin{table*}[htb]
\centering
\begin{small}
\setlength{\tabcolsep}{1.8mm}{
\begin{tabular}{lccccccc}
\toprule
    \textbf{Method} &\textbf{PickScore $\uparrow$} & \textbf{Aes. Score $\uparrow$} & \textbf{HPS $\uparrow$}& \textbf{ CLIP-S $\uparrow$} & \textbf{ D-CLIP-S $\uparrow$}& \textbf{CI Score $\uparrow$}& \textbf{Avg. Rank. $\downarrow$} \\
\hline
ChatGPT & 19.338
& 6.145
& 20.123
& \textbf{28.837}
& 14.594
& \textbf{87.496} & \underline{2.500} \\
\hline
InstructPix2Pix&  19.235
& 5.917
& 19.430
& 24.580
& \textbf{20.818} 
& -  & 3.400 \\
\hline
\emph{DiffChat} (SFT only) & \underline{19.339}
& \underline{6.149}
& \underline{20.129}
& 28.769
& 15.532
& 85.089& \underline{2.500}  \\
\emph{DiffChat} (full imp.)  & \textbf{19.359}
& \textbf{6.169}
& \textbf{20.163}
& \underline{28.822}
& \underline{15.747}
& \underline{87.314} & \textbf{1.500} \\
\bottomrule
\end{tabular}}
\end{small}\\
(a) Stable Diffusion 1.5 \bigskip

\begin{small}
\setlength{\tabcolsep}{1.8mm}{
\begin{tabular}{lccccccc}
\toprule
    \textbf{Method} &\textbf{PickScore $\uparrow$} & \textbf{Aes. Score $\uparrow$} & \textbf{HPS $\uparrow$}& \textbf{ CLIP-S $\uparrow$} & \textbf{ D-CLIP-S $\uparrow$}& \textbf{CI Score $\uparrow$} & \textbf{Avg. Rank. $\downarrow$} \\
\hline
ChatGPT & \underline{19.602}
& \underline{6.501}
& 21.010
& \underline{29.692}
& 16.863
& \textbf{87.496} & \underline{2.333} \\
\hline
InstructPix2Pix&  19.349
& 5.977
& 19.672
& 25.175
& \textbf{19.309}
& - & 3.400 \\
\hline
\emph{DiffChat} (SFT only) & \underline{19.602}
& 6.487
& \textbf{21.042}
& 29.686
& 17.592
& 85.089 & 2.500
 \\
\emph{DiffChat} (full imp.)  & \textbf{19.612}
& \textbf{6.511}
& \underline{21.032}
& \textbf{29.723}
& \underline{17.687}
& \underline{87.314} & \textbf{1.500}
  \\
\bottomrule
\end{tabular}}
\end{small}
\\(b) Dreamlike \bigskip

\begin{small}
\setlength{\tabcolsep}{1.8mm}{
\begin{tabular}{lccccccc}
\toprule
    \textbf{Method} &\textbf{PickScore $\uparrow$} & \textbf{Aes. Score $\uparrow$} & \textbf{HPS $\uparrow$}& \textbf{ CLIP-S $\uparrow$} & \textbf{ D-CLIP-S $\uparrow$}& \textbf{CI Score $\uparrow$} & \textbf{Avg. Rank. $\downarrow$} \\
\hline
ChatGPT & 19.494
& \underline{6.254}
& 20.586
& \underline{28.863}
& 15.258
& \textbf{87.496} & \underline{2.500} \\
\hline
InstructPix2Pix&  19.313
& 6.052
& 19.784
& 25.072
& \textbf{20.790}
& - & 3.400 \\
\hline
\emph{DiffChat} (SFT only) & \underline{19.500}
& 6.249
& \underline{20.589}
& 28.857
& 16.018
& 85.089 & 2.667
 \\
\emph{DiffChat} (full imp.)  & \textbf{19.513}
& \textbf{6.283}
& \textbf{20.622}
& \textbf{28.945}
& \underline{16.242}
& \underline{87.314} & \textbf{1.333} \\
\bottomrule
\end{tabular}}
\end{small}
\\(c) Realistic \bigskip

\begin{small}
\setlength{\tabcolsep}{1.8mm}{
\begin{tabular}{lccccccc}
\toprule
    \textbf{Method} &\textbf{PickScore $\uparrow$} & \textbf{Aes. Score $\uparrow$} & \textbf{HPS $\uparrow$}& \textbf{ CLIP-S $\uparrow$} & \textbf{ D-CLIP-S $\uparrow$}& \textbf{CI Score $\uparrow$} & \textbf{Avg. Rank. $\downarrow$}\\
\hline
ChatGPT & \underline{19.590}
& \underline{6.312}
& 20.755
& 28.982
& 14.994
& \textbf{87.496} & \underline{2.500} \\
\hline
InstructPix2Pix&  19.363
& 6.082
& 19.913
& 25.209
& \textbf{20.602}
& - & 3.400 \\
\hline
\emph{DiffChat} (SFT only) & \underline{19.590}
& 6.311
& \underline{20.760}
& \underline{29.051}
& 15.842
& 85.089 & \underline{2.500} \\
\emph{DiffChat} (full imp.)  & \textbf{19.600}
& \textbf{6.335}
& \textbf{20.780}
& \textbf{29.101}
& \underline{16.214}
& \underline{87.314} & \textbf{1.333}  \\
\bottomrule
\end{tabular}}
\end{small}
\\(d) Deliberate \bigskip

\begin{small}
\setlength{\tabcolsep}{1.8mm}{
\begin{tabular}{lccccccc}
\toprule
    \textbf{Method} &\textbf{PickScore $\uparrow$} & \textbf{Aes. Score $\uparrow$} & \textbf{HPS $\uparrow$}& \textbf{ CLIP-S $\uparrow$} & \textbf{ D-CLIP-S $\uparrow$}& \textbf{CI Score $\uparrow$} & \textbf{Avg. Rank. $\downarrow$}\\
\hline
ChatGPT & 19.820
& 6.757
& 21.636
& \textbf{30.395}
& 17.300
& \textbf{87.496} & 2.333 \\
\hline
InstructPix2Pix&  19.468
& 4.603
& 16.381
& 13.313
& 5.117
& - & 4.000 \\
\hline
\emph{DiffChat} (SFT only) & \underline{19.823}
& \underline{6.762}
& \textbf{21.659}
& 30.383
& \underline{17.992}
& 85.089 & \underline{2.167} \\
\emph{DiffChat} (full imp.)  & \textbf{19.836}
& \textbf{6.781}
& \underline{21.654}
& \underline{30.392}
& \textbf{18.040}
& \underline{87.314} & \textbf{1.500} \\
\bottomrule
\end{tabular}}
\end{small}
\\(e) Stable Diffusion XL 1.0
\caption{Detailed automatic evaluation results on the~\textbf{InstructPE} testing set with different SD models. 
Avg. Rank. is calculated as the average ranking value under each score. Aes. Score: the aesthetic score.
CLIP-S: CLIP score.
D-CLIP-S: directional CLIP similarity. 
SFT only: only conducting supervised fine-tuning.
Full imp.: Full implementation.
}
\label{detail_result}
\end{table*}

\subsection{Collaborating with various TIS models}\label{multiple-versions}
\begin{figure*}[h]
\centering
\includegraphics[width=0.98\textwidth]{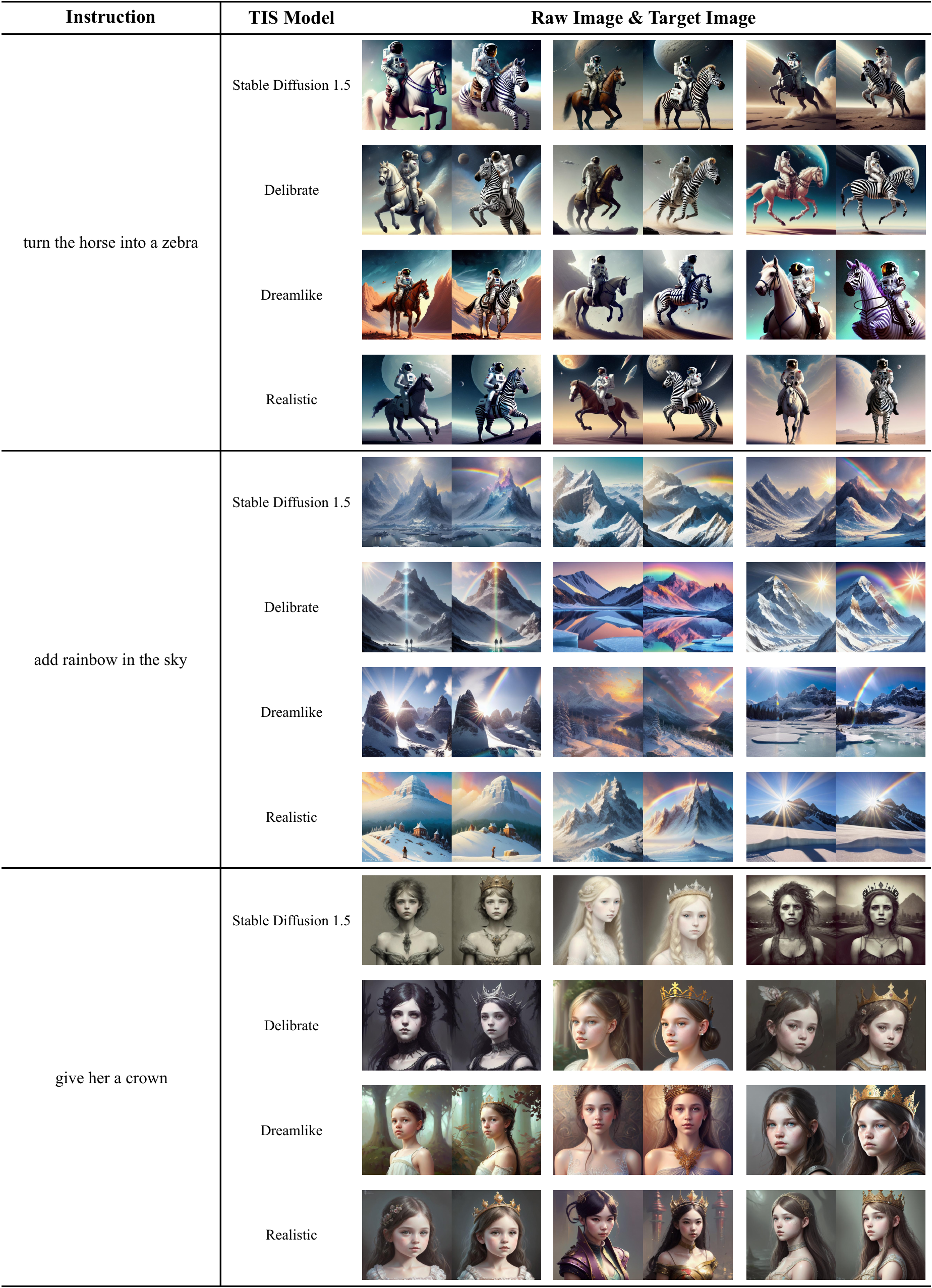}
\caption{Examples of raw images \& target images generated with collaboration between \textit{DiffChat} and   various Stable Diffusion-style models.} 
\label{versions}
\end{figure*}

Fig.~\ref{versions} shows some examples of raw images and target images generated with collaboration between \textit{DiffChat} and various
Stable Diffusion-style models: Deliberate, Dreamlike, and Realistic.
The transferability of \textit{DiffChat} is verified.

\end{document}